%% file: iclr2023_conference.tex
\documentclass{article} 
\usepackage{iclr2023_conference,times}

\input{math_commands.tex}

\usepackage[colorlinks=true,linkcolor=red,anchorcolor=red,citecolor=blue,filecolor=red,menucolor=red,runcolor=red,urlcolor=red]{hyperref}
\usepackage{url}
\usepackage{booktabs}       
\usepackage[noabbrev,capitalize]{cleveref}  
\usepackage[noabbrev,capitalize]{cleveref}  
\usepackage[affil-it]{authblk}

\usepackage[english]{babel}
\usepackage{blindtext}

\usepackage{color, colortbl}
\usepackage{graphicx}
\definecolor{Gray}{gray}{0.93}

\newcommand{\ie}{\textit{i.e.}}
\newcommand{\eg}{\textit{e.g.}}

\title{Towards Federated Learning Under Resource Constraints via Layer-wise Training\\ and Depth Dropout}

\author[1]{Pengfei Guo\thanks{Work done during an internship at Google Research. } \,}
\author[2]{Warren Richard Morningstar}
\author[2]{Raviteja Vemulapalli}
\author[2]{Karan Singhal}
\author[1]{Vishal M. Patel}
\author[2]{Philip Andrew Mansfield}
\affil[1]{Johns Hopkins University}
\affil[2]{Google Research}
\affil[ ]{\texttt {\{pguo4,vpatel36\}@jhu.edu}, {\texttt{\{wmorning,ravitejavemu,karansinghal,memes\}@google.com}}}

\iclrfinalcopy 
\begin{document}

\maketitle

\input{p1_abstract}
\input{p2_introduction}

\input{p3_method}
\input{p4_experiment}

\input{p5_conclusion}


\input{iclr2023_conference.bbl}
\bibliographystyle{iclr2023_conference}

\end{document}

%% file: math_commands.tex
\usepackage{amsmath,amsfonts,bm}

\def\eqref#1{equation~\ref{#1}}

\def\1{\bm{1}}

\DeclareMathAlphabet{\mathsfit}{\encodingdefault}{\sfdefault}{m}{sl}
\SetMathAlphabet{\mathsfit}{bold}{\encodingdefault}{\sfdefault}{bx}{n}



%% file: p1_abstract.tex
\begin{abstract}
Large machine learning models trained on diverse data have recently seen unprecedented success. Federated learning enables training on private data that may otherwise be inaccessible, such as domain-specific datasets decentralized across many clients. However, federated learning can be difficult to scale to large models when clients have limited resources. This challenge often results in a trade-off between model size and access to diverse data. To mitigate this issue and facilitate training of large models on edge devices, we introduce a simple yet effective strategy, \textit{Federated Layer-wise Learning}, to simultaneously reduce per-client memory, computation, and communication costs.
Clients train just a single layer each round, reducing resource costs considerably with minimal performance degradation. We also introduce \textit{Federated Depth Dropout}, a complementary technique that randomly drops frozen layers during training, to further reduce resource usage. Coupling these two techniques enables us to effectively train significantly larger models on edge devices. Specifically, we reduce training memory usage by 5$\times$ or more in federated self-supervised representation learning, and demonstrate that performance in downstream tasks is comparable to conventional federated self-supervised learning.  
\end{abstract}

%% file: p2_introduction.tex
\section{Introduction}\label{sec:intro}

Over the last several years, deep learning has witnessed a rapid paradigm shift towards large foundational models trained on massive datasets \citep{brown2020language,chowdhery2022palm}. These models learn representations which often extend to diverse downstream tasks. However, when pre-training data is distributed across a large number of devices, it becomes impractical to train models using centralized learning. In these cases, Federated Learning \citep[FL;][]{konevcny2016federated} allows participating clients to train a model together without exchanging raw data. This privacy-preserving property makes FL a popular choice for a range of applications, including face recognition~\citep{mei2022escaping}, autonomous driving~\citep{li2021privacy}, recommendation systems~\citep{ning2021learning}, and self-supervised representation learning~\citep{vemulapalli2022federated}. In self-supervised learning, SimCLR~\citep{chen2020simple}, BYOL~\citep{grill2020bootstrap}, and SimSiam~\citep{chen2021exploring} are widely used approaches that can be adapted for use in FL settings using algorithms like Federated Averaging \citep[FedAvg;][]{mcmahan2017communication}. Representation learning benefits from large models due to their capacity to learn more nuanced and reliable representations of the data \citep{chen2022pali,tran2022plex}. However, in cross-device FL settings, the limited resources of edge devices (including memory, computation capacity, and network bandwidth) impedes the development of large models \citep{wang2021field,FedGuide}. In this work, we focus on federated training of large representation learning models on a large number of edge devices under resource constraints.


Typically in FL, clients' models share a single global architecture and perform end-to-end training in each communication round \citep{mcmahan2017communication}. However, many edge devices (e.g., Internet of Things (IoT) devices, mobile phones, tablets, and personal computers) lack sufficient memory and compute to train most existing large ML models. For example, the Google Pixel 6 has 12 GB of memory, which is insufficient to naively train a multi-billion parameter model. Communication of such a model and its gradient updates during every round of FL is also prohibitively data-intensive and time-consuming. These resource constraints create obstacles for real-world federated learning applications with large-scale models. 

\paragraph{Related Work} One direction to manage resource constraints for federated learning on edge devices is to carefully select model architecture and hyperparameters~\citep{cheng2022does} to ensure that it can be trained and run efficiently on edge devices. Another direction is to use techniques such as model compression~\citep{xu2020ternary} and pruning~\citep{jiang2022model} to reduce the size and complexity of the model, making it more suited for training and deployment on edge devices. This can be done by removing redundant or unnecessary layers within the model, or by using low-precision arithmetic to reduce the amount of memory and computation required. In both cases, model performance degradation is usually unavoidable. Some methods rely on partially local models to avoid communicating entire models with a central server \citep{singhal2021federated}, but these approaches do not reduce local memory usage on edge devices. Other approaches involve retaining part of a model on a central server \citep{augenstein2022mixed}, which can reduce the need for local resource usage and is thus complementary to our work but does not itself enable training larger local models. In addition, FL methods designed for heterogeneous systems~\citep{caldas2018expanding,horvath2021fjord,mei2022resource} are able to construct sub-models at different complexities from one unified base model. One of the early works in this direction is Federated Dropout~\citep{caldas2018expanding}, which allows users to train using smaller subsets of the global model, reducing the client communication and computation costs. Empirically, this method can drop up to 50\% of model parameters, but will degrade model performance. FjORD~\citep{horvath2021fjord} improves upon Federated Dropout by introducing an ordered dropout technique that drops adjacent components of the model instead of random neurons. Experiments by \citet{horvath2021fjord} show that ordered dropout can bring computational benefits and better model performance. More recently, FLANC~\citep{mei2022resource} formulates networks at different capacities as linear combinations of a shared neural basis set, so sub-models can be composed by using capacity-specific coefficients. While these methods can reduce average local resource usage in FL, full model training is still needed for certain clients, and thus the resource usage upper bound is still determined by the base model size. 

\paragraph{Our Contributions} Full utilization of available resources in cross-device FL remains a challenging task. In this paper, we propose \emph{Federated Layer-wise Learning}, a strategy for
resource-saving federated training. 
In particular, training is divided into several phases. In each phase, we update only one active layer and freeze parameters in fixed layers. 
As shown in Fig.~\ref{fig:fig2}(b), our experimental evaluation demonstrates that Federated Layer-wise Learning (FLL) can significantly reduce the resource usage of a single client compared to federated end-to-end learning (FEL) in all aspects. Specifically, FLL only uses 7--22\% memory, 8--39\% computation, and 8--54\% communication compared to FEL. In addition, we demonstrate that \emph{Depth Dropout} is an effective complementary strategy in federated layer-wise learning, which further reduces resource usage upper bounds without degrading model performance.


%% file: p3_method.tex
\section{Methods}
\begin{figure*}[t] 
\centering
\includegraphics[width=\linewidth]{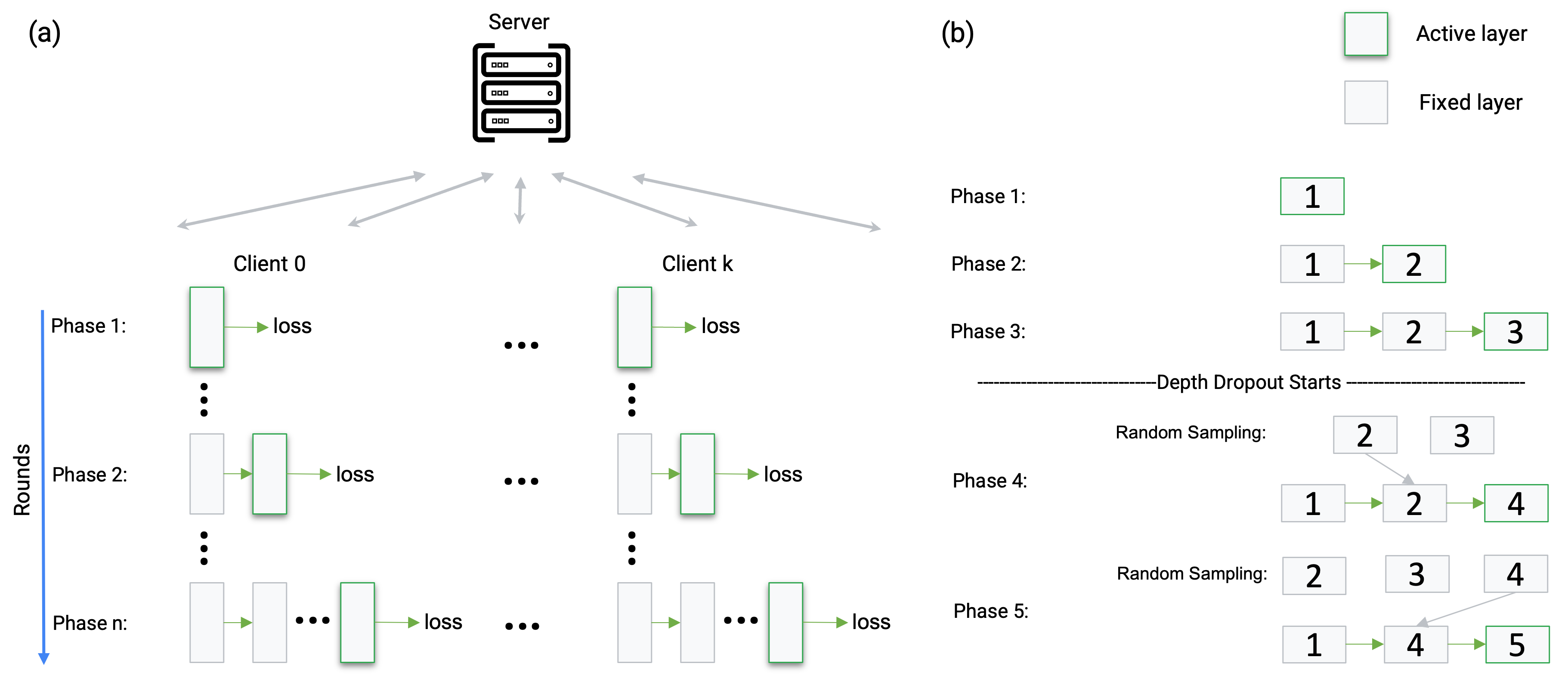}
\vskip-10pt
\caption{(a) Overview of Federated Layer-wise Learning. (b) Schematics of training procedure with Depth Dropout for a 5-layer model with a budget of 3 layers.}
\label{fig:overview}
\end{figure*}

We consider the canonical cross-device FL scenario, in which a large distributed population of clients contributes to training a global model using their locally stored data \citep{FedGuide}.  In such scenarios, the general training process involves the following steps:  first, a coordinating server sends the current set of model parameters to each contributing device. Next, each device runs a local training algorithm and sends the result back to the server. Finally, the server aggregates the model updates received from all devices to determine the new set of model parameters and restarts the cycle. As previously discussed, device resource constraints limit real-world large-scale federated learning applications and lead to a trade-off between model complexity and data accessibility. 

\subsection{Federated Layer-wise Learning}


To address resource constraints when training models in FL, we propose a simple yet effective Federated Layer-wise Learning technique.  We motivate and apply the method to self-supervised learning~\citep{chen2020simple} in this work, but this approach is also broadly applicable. In contrast to downstream vision tasks (\eg, classification) that require the extraction of compact features (\ie, the interpretation of input) from the output of neural networks, contrastive representation learning employs the principle of learning representations of the input data by comparing and contrasting it with other similar and dissimilar examples. Since this loss only refers to layer activations, it can be attached to any encoder layer. When applied to residual networks, we expect the effect of applying the loss on successive layers to be progressive. This motivates our Federated Layer-wise Learning method, as depicted in Fig.~\ref{fig:overview}(a). 

The proposed method divides the holistic training process into several phases and progressively grows the model in an incremental schedule, starting from the shallow layers and moving to deeper layers. Each layer is trained for a predefined number of communication rounds before proceeding to the next layer. We only need to compute gradients and upload them to the server for the active layer, which simultaneously reduces memory usage, compute, and communication costs. We can control resource usage by varying the number of active and fixed layers during training, potentially treating multiple layers as active at a given round. As an aside, in cross-silo FL (where clients participate repeatedly in training) the fact that only one active layer is being trained enables us to avoid communicating the rest of the model to devices on most rounds, further reducing communication.

\subsection{Depth Dropout}

While the proposed Federated Layer-wise Learning significantly alleviates resource usage, our target scenario is cross-device FL, in which only a relatively small subset of active clients are selected from a large pool of participants. It is likely that a given client will not be selected twice during the entire federated training process \citep{wang2021field}. Thus, it is necessary to download both fixed and active layers from the server to the clients. This can still present a challenge for clients with resource constraints, as downloading a large number of fixed layers and performing forward passes can be computationally intensive at the end of the training process. To this end, we propose \emph{Depth Dropout} to address the increasing resource usage introduced by a large number of fixed layers. Fig.~\ref{fig:overview}(b) shows how to apply Depth Dropout to a 5-layer model with a budget of 3 layers. It begins by progressively expanding the model to reach its maximum capacity, which in this case is 3 layers. During the initial three phases, we perform standard layer-wise training. In the last two phases, we randomly remove certain fixed layers. However, the first layer, which includes Transformer patch encoding and position embedding, is never removed. For example, in phase 4, we have the option to remove either layer 1 or layer 2, while in phase 5, we have three candidates to remove. This randomization process is akin to the Dropout technique used in neural networks, and is only applied during training. During inference, the full model with 5 layers is utilized.

%% file: p4_experiment.tex
\section{Experiments}\label{sec:exp}

\paragraph{Datasets and Implementation} We partition the standard CIFAR-100~\citep{krizhevsky2009learning} training set into 125 clients to simulate a cross-device FL scenario. The original test set in CIFAR-100~\citep{krizhevsky2009learning} is considered the global test set used to measure performance. ViT-Ti/16~\citep{dosovitskiy2020image} is used as the representation learning backbone. All models are trained using the following settings: SGD optimizer for the server and clients; client learning rate of $1\times10^{-3}$; batch size of 16; 32 active clients per round. 

\paragraph{The Effectiveness of Layer-wise Training} Here we compare our approach with federated end-to-end learning on standard benchmarks. Results with different setups on CIFAR-100 are shown in Table~\ref{tab:main}. We can make the following observations: (i) Both models pre-trained by Federated Layer-wise Learning and federated end-to-end learning can significantly outperform the model without pre-training, indicating the effectiveness of self-supervised representation learning in federated settings.
(ii) While the Federated Layer-wise Learning approach is an approximation of federated end-to-end learning, it can achieve performance on par with the end-to-end method in downstream evaluation tasks. In particular, the performance gap is less than 1\% when using the representation from the last layer (layer 12) of the network. (iii) we found that intermediate representations from the Federated Layer-wise Learning model performed better than those from the federated end-to-end learning model in certain downstream tasks. For example, in a linear downstream task using the representation from layer 3, the Federated Layer-wise Learning model achieved 28.3\% accuracy, while the federated end-to-end learning model achieved 23.9\%. This trend was also observed in other downstream tasks using different intermediate representations. This superior performance of intermediate representations is due to the contrastive loss being applied to all layers during the layer-wise pre-training process. These results suggest that models trained using the proposed method can easily compose sub-models of varying complexities.

\begin{figure*}[t!] 
\centering
\includegraphics[width=\linewidth]{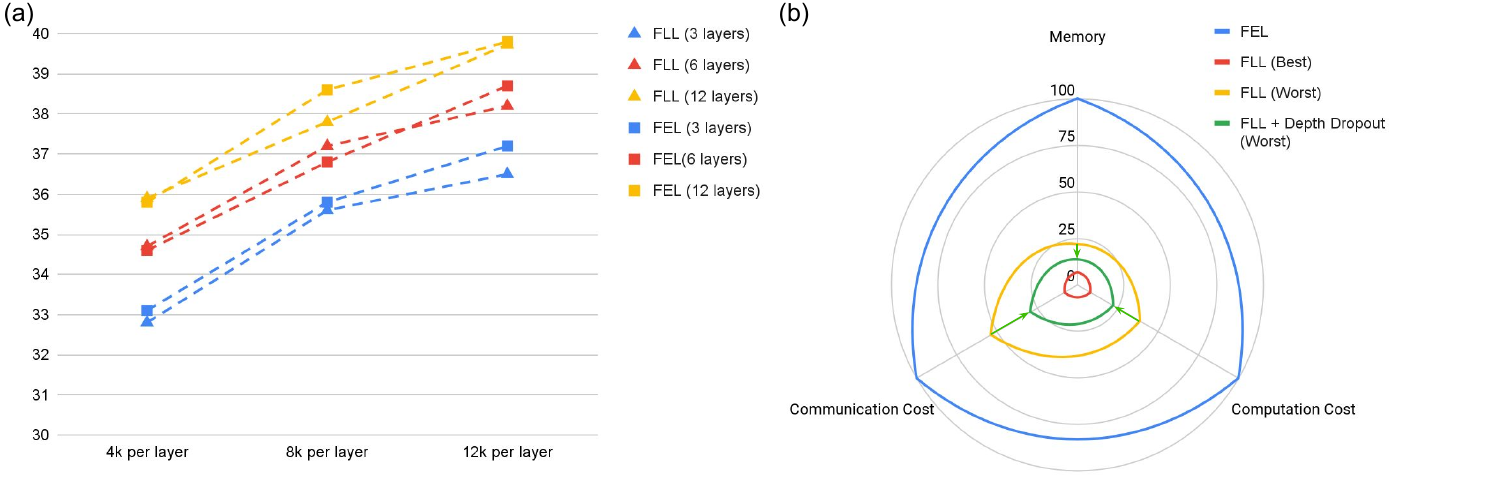}
\vskip-10pt
\caption{(a) Comparison between Federated Layer-wise Learning (FLL) and federated end-to-end learning (FEL) under different numbers of training rounds and model sizes. Results are
reported on CIFAR-100 with downstream finetuning evaluation. (b) Resource usage comparison of a client per-round after combining Federated Layer-wise Learning and Depth Dropout. }
\label{fig:fig2}
\end{figure*}

We conducted additional experiments to further investigate the effect of model size (number of layers) and number of training rounds per layer on the performance of Federated Layer-wise and end-to-end learning. The results of these experiments are shown in Figure~\ref{fig:fig2}(a). Increasing the number of layers generally led to improved performance for both learning approaches. We also found that the difference in performance between the two approaches was minimal when the number of training rounds per layer was small (4k) but became more pronounced when the number of training rounds per layer was increased (12k). Based on these results, it appears that layer-wise learning may require slightly more training rounds per layer to reach the same performance as end-to-end learning. This may be due to the fact that layer-wise learning is an approximation of end-to-end learning. However, the performance gap between the two approaches is generally less than 1\%.

\input{tables/main_table}
\input{tables/depth_dropout}
\noindent\textbf{The Effectiveness of Depth Dropout.} We evaluate Depth Dropout with Federated Layer-wise Learning. We conducted two sets of experiments: applying Depth Dropout to a 6-layer model and a 12-layer model, with a fixed dropout rate of 50\% (meaning half of the fixed layers were dropped). The results, shown in \cref{tab:ddrop-finetuning,tab:ddrop-linear}, demonstrate that Depth Dropout does not significantly impact model performance. For example, the 6-layer model with Depth Dropout achieved 37.0\% accuracy after finetuning, while the 6-layer model trained with only Layer-wise Learning achieved 37.2\% accuracy. We observed similar results for the 12-layer model with Depth Dropout, which achieved 37.6\% accuracy after finetuning, compared to 37.8\% for the model trained with normal Layer-wise Learning. Additionally, Depth Dropout significantly reduced resource usage. It is worth noting that the resource usage of the 12-layer model with a budget of 6 layers was equivalent to the resource usage of a 6-layer model without Depth Dropout. As shown in Fig.~\ref{fig:fig2}, depth dropout reduced the upper bounds of resource usage in all three categories, especially communication cost. The original upper bound for Layer-wise training was 54\%, but it was reduced to 29\% when the dropout rate was set to 50\%.

%% file: tables/main_table.tex
\begin{table}[ht!]
\caption{Experimental results on CIFAR-100 with different pre-training strategies. For image classification, we report standard Top-1 accuracy (\%). }
\setlength{\tabcolsep}{2.5pt}
\label{tab:main}
\small
\centering
\begin{tabular}{ccccccccc}
\toprule\rowcolor{white}
\multicolumn{1}{c|}{Downstream} & \multicolumn{4}{c|}{Linear}                                 & \multicolumn{4}{c}{Finetune}           \\ 
\multicolumn{1}{c|}{Representation From} & Layer 1 & Layer 3 & Layer 6 & \multicolumn{1}{c|}{Layer 12} & Layer 1 & Layer 3 & Layer 6 & Layer 12 \\ \midrule\rowcolor{white}
Pre-training Method & \multicolumn{8}{c}{Federated Layer-wise  Learning}                                                   \\ \hline
Accuracy            & 25.3    & 28.3    & 29.2    & \multicolumn{1}{c}{29.8}     & 30.1    & 35.6    & 37.2    & 37.8     \\ \hline\rowcolor{white}
Pre-training Method & \multicolumn{8}{c}{Federated End-to-end Learning}                                                    \\ \hline
Accuracy            & 18.0    & 23.9    & 27.8    & \multicolumn{1}{c}{30.3}     & 25.4    & 32.2    & 35.7    & 38.6     \\ \hline\rowcolor{white}
Pre-training Method & \multicolumn{8}{c}{Training from scratch (Without Pre-training)}                                     \\ \hline
Accuracy            & 9.2     & 10.0    & 10.5    & \multicolumn{1}{c}{11.3}     & 18.7    & 24.2    & 27.6    & 29.2     \\ \bottomrule
\end{tabular}
\end{table}

%% file: tables/depth_dropout.tex
\begin{table}[ht!]
\caption{Accuracy of depth dropout with Federated Layer-wise Learning, under finetuning downstream evaluation. The budget specifies the max number of layers involved in training.  }
\setlength{\tabcolsep}{4.0pt}
\label{tab:ddrop-finetuning}
\centering
\small
\begin{tabular}{ccccccc}
\toprule\rowcolor{white}
Model Size          & 6 layers  & \begin{tabular}[c]{@{}c@{}}6 layers \\ (Budget: 3 layers)\end{tabular} & 3 layers & 12 layers & \begin{tabular}[c]{@{}c@{}}12 layers\\ (Budget: 6 layers)\end{tabular} & 6 layers  \\ \midrule\rowcolor{white}
Accuracy            & 37.2      & 37.0                                                                   & 32.8     & 37.8      & 37.6                                                                   & 37.2  \\  \bottomrule
\end{tabular}
\end{table}

\begin{table}[ht!]
\caption{Accuracy of depth dropout with Federated Layer-wise Learning, under linear downstream evaluation. The budget specifies the max number of layers involved in training. }
\setlength{\tabcolsep}{4.0pt}
\label{tab:ddrop-linear}
\centering
\small
\begin{tabular}{ccccccc}
\toprule\rowcolor{white}
Model Size          & 6 layers  & \begin{tabular}[c]{@{}c@{}}6 layers \\ (Budget: 3 layers)\end{tabular} & 3 layers & 12 layers & \begin{tabular}[c]{@{}c@{}}12 layers\\ (Budget: 6 layers)\end{tabular} & 6 layers  \\ \midrule\rowcolor{white}
Accuracy            & 29.2      & 29.1                                                                   & 28.3     & 29.8      & 29.7                                                                   & 29.2  \\  \bottomrule
\end{tabular}
\end{table}


%% file: p5_conclusion.tex
\section{Conclusion}\label{sec:conc}

Our study presents Federated Layer-wise Learning for devices with limited resources, which simultaneously reduces the demands on memory, computation, and communication for individual clients without significantly compromising performance in comparison to end-to-end training. We demonstrate that our proposed Depth Dropout technique is an effective complement to Federated Layer-wise Learning, as it further reduces resource usage across all categories with minimal loss of performance, even when dropping half of the fixed layers. Future work can evaluate these methods on larger-scale and naturally partitioned datasets, which would enable more realistic analysis of generalization performance across devices~\citep{yuan2021we}. Additionally, we intend to investigate the effects of varying dropout rate for the Depth Dropout technique. Furthermore, our method can be integrated with other memory-efficient training techniques, such as model compression~\citep{deng2020model} and activation paging~\citep{patil2022poet} to potentially further reduce resource usage.